\documentclass[letterpaper]{article}
\usepackage{uai2020}
\usepackage[margin=1in]{geometry}

\usepackage[round]{natbib}
\usepackage{amssymb}
\usepackage{amsmath}
\usepackage{multirow}
\usepackage{subcaption}
\usepackage{graphicx}
\usepackage{booktabs}       
\usepackage[T1]{fontenc}    

\usepackage{times}

\title{The Effect of Optimization Methods on the Robustness\\ of Out-of-Distribution Detection Approaches}

\author{ {\bf Vahdat Abdelzad\thanks{Department of Electrical and Computer Engineering, University of Waterloo}} \\
vabdelza@uwaterloo.ca \\
\And
{\bf Krzysztof Czarnecki\textsuperscript{\rm *}}  \\
kczarnec@gsd.uwaterloo.ca\\
\And
{\bf Rick Salay\textsuperscript{\rm *}}   \\
rsalay@gsd.uwaterloo.ca\\
}

\begin{document}

\maketitle

\begin{abstract}
Deep neural networks (DNNs) have become the de facto learning mechanism in different domains. Their tendency to perform unreliably on out-of-distribution (OOD) inputs hinders their adoption in critical domains. Several approaches have been proposed for detecting OOD inputs. However, existing approaches still lack robustness. In this paper, we shed light on the robustness of OOD detection (OODD) approaches by revealing the important role of optimization methods. We show that OODD approaches are sensitive to the type of optimization method used during training deep models. Optimization methods can provide different solutions to a non-convex problem and so these solutions may or may not satisfy the assumptions (e.g., distributions of deep features) made by OODD approaches. Furthermore, we propose a robustness score that takes into account the role of optimization methods. This provides a sound way to compare OODD approaches. In addition to comparing several OODD approaches using our proposed robustness score, we demonstrate that some optimization methods provide better solutions for OODD approaches.
\end{abstract}

\section{INTRODUCTION} \label{intoduction}

Deep neural networks (DNNs) are considered to be the state-of-the-art learning method used in domains such as automated driving and medical diagnoses. Despite their excellent performance, they are vulnerable to out-of-distribution (OOD) inputs \citep{deep_fool_2014, Hendrycks-baseline-2017}. Several approaches have been proposed to detect OOD inputs~\citep{Hendrycks-baseline-2017, devries2018learning, Liang-ODIN-2018, Lee-mah-2019}.
The quality of these OODD approaches is evaluated based on their OODD performance for several OOD datasets relative to a specific model trained for an in-distribution (ID) dataset. However, we can observe cases in which some OODD approaches do not perform according to their claims and in some cases even worse than baselines \citep{Lee-mah-2019,ren_ratio_2019}.

For example, ODIN \citep{Liang-ODIN-2018} is deemed to be better than max-softmax as a baseline \citep{Hendrycks-baseline-2017}, but \cite{Lee-mah-2019}'s experiments show that in some cases ODIN is worse than max-softmax. In the same vein, \cite{ren_ratio_2019} show that the Mahalanobis Distance (MD) approach proposed by \cite{Lee-mah-2019}, which is supposed to be better than ODIN and max-softmax, can be worse than max-softmax. These observations make it challenging to trust an OODD approach or compare OODD approaches with each other.

In this paper, we shed light on aspects that have been overlooked during the evaluation of OODD approaches, in particular, the role of optimization methods. Deep models, in our context classification models, can have the same classification accuracy over the test set under different optimization methods. These models are the same from the perspective of classification accuracy, but they might converge to different local minima (due to the non-convex nature of the problem). Therefore, they offer distinct solutions to the same problem.

We show that OODD approaches can be sensitive to these different solutions that exist for the given problem (i.e., image classification). In other words, OODD approaches can have different detection performance for distinct solutions offered by optimization methods to the same problem. Therefore, comparing OODD approaches with each other over a solution offered by one optimization method can not provide a valid comparison because one OODD approach can be better than others for that specific solution and vice versa. We propose a robustness score tailored to take into account this aspect of detection. Our robustness score is approximated separately for each OODD metric and provides a sound way of comparison for OODD approaches. We compare several OODD approaches (over different OODD metrics) based on our robustness score. We also demonstrate that some optimization methods provide better solutions for detecting OOD inputs; thus, they are better candidates for training DNNs in safety-critical systems.

\begin{table}
\centering
\caption{Test classification accuracy of a CNN for Fashion-MNIST under different optimization methods.}
    \begin{tabular}{lc}
        \toprule
        Optimizer & accuracy \\
        \bottomrule
        
        Adam      & \multicolumn{1}{l}{\textit{0.9704}}  \\
        RMSprop   & \multicolumn{1}{l}{\textit{0.9698}} \\
        Adamax    & \multicolumn{1}{l}{\textit{0.9560}} \\
        Nadam     & \multicolumn{1}{l}{\textit{0.9612}}\\ 
        SGD       & \multicolumn{1}{l}{\textit{0.9705}} \\
        Adagrad   & \multicolumn{1}{l}{\textit{0.9650}} \\
        Adadelta  & \multicolumn{1}{l}{\textit{0.9707}}\\
        
    \end{tabular}
    \label{table_fashion-mnist_try02_accuracies}
\end{table}
\begin{figure}
    \centering
    \includegraphics[width=7.5cm]{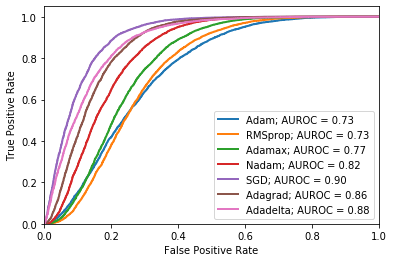}
    \caption{AUROC (based on the MD approach) of models in Table ~\ref{table_fashion-mnist_try02_accuracies} when the OOD dataset is uniform noise. }  
    \label{fashion_mnist_accuracy_md_unifrom}
\end{figure}
\begin{figure*}
    \centering
    \subcaptionbox{\label{fashion-mnist_md_max-softmax_optimizers_comp:ma}}{\includegraphics[width=6cm]{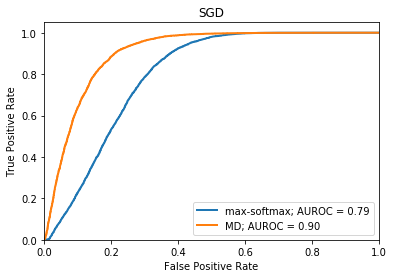}}
    \subcaptionbox{\label{fashion-mnist_md_max-softmax_optimizers_comp:mb}}{\includegraphics[width=6cm]{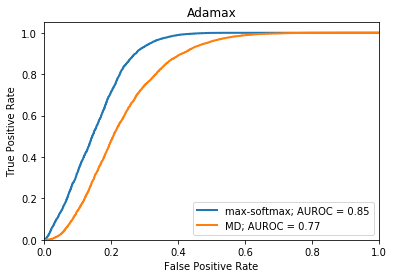}}
    
    \caption{AUROC for detecting uniform noise (as OOD) by max-softmax and MD over the test set of Fashion-MNIST (as ID). a) It shows AUROC for these two OODD approaches under the SGD optimizer. b) It shows AUROC for these two approaches under the Adamax optimizer. As can be seen, there is not an overall winner. }  
    \label{fashion-mnist_md_max-softmax_optimizers_comp}
\end{figure*}

\section{THE PROBLEM STATEMENT} \label{problem_statement}
Stochastic gradient descent (SGD) is one of the basic methods used for optimizing objective functions of DNNs. Although it guarantees to converge in convex settings, it is slow. Adaptive gradient methods such as Adam \citep{Adam_Kingma_2015} have been introduced to reduce the  training time. In non-convex settings (e.g., image classification), optimization methods can cause deep models to converge to different local minima. When the classification accuracy of these models is measured based on a test dataset, they can have almost the same accuracy. In this case, it is generally accepted that any of these models can be deployed in real life. 

However, these models are different internally. For DNNs, it means weights of these models vary and so they focus on different features to obtain their posteriors. Therefore, an OODD approach based on posteriors or deep features may demonstrate different OODD performance for these models. When this phenomenon happens for an OODD approach, it implies that the approach is not robust to the choice of optimization method.

Table \ref{table_fashion-mnist_try02_accuracies} shows the classification accuracy of a convolutional neural network (CNN) trained on the Fashion-MNIST dataset with several well-known optimization methods. As can be seen, they all converge and have similar classification accuracy. Figure \ref{fashion_mnist_accuracy_md_unifrom} shows the Area Under the Receiver Operating Characteristic curve (AUROC) for the uniform noise (as an OOD dataset) based on the MD approach (over logits features) for models in Table \ref{table_fashion-mnist_try02_accuracies}. There is a large variance for AUROC values even though the models have similar classification accuracy. Please note that the differences in Figure \ref{fashion_mnist_accuracy_md_unifrom} are large, relatively to the typical differences between different OODD approaches. A similar case for MNIST as the ID dataset is depicted in the Appendix, Figure  ~\ref{mnist_accuracy_md_unifrom}.

This characteristic of OODD approaches makes them incomparable unless the effect of the optimization method is incorporated into the comparison. Figure \ref{fashion-mnist_md_max-softmax_optimizers_comp:ma} shows a comparison of AUROCs calculated by two OODD approaches, max-softmax and MD, under the SGD optimization method. As can be seen, the MD approach has much better AUROC than max-softmax. Figure \ref{fashion-mnist_md_max-softmax_optimizers_comp:mb} shows AUROCs under Adamax and in which max-softmax has better AUROC than MD. The classification accuracies based on the SGD and Adamax optimization methods are 0.9705 and 0.956, respectively. Thus, the best AUROC is not necessarily related to the best performing classifier when max-softmax is used.

This sensitivity to optimization method may be a possible reason for discrepancies seen among comparison results of published articles in this area. It also highlights the fact that an OODD approach should not be applied blindly to deep models based on its detection performance over a specific ID, OOD, and optimization method. Furthermore, this phenomenon is not limited to AUROC---it can also be observed when OODD approaches are compared using other common metrics as defined in Section \ref{evaluation-metrics}.

In order to incorporate this sensitivity of OODD approaches into OODD metrics and possibly find optimization methods that provide better solutions for OODD approaches, we propose a robustness score that takes into account both the level of detection performance and variation due to different optimization methods. It is described in details in the following section.

\section{PROPOSED ROBUSTNESS SCORE}
In this section, we formulate our proposed robustness score. We first formulate a robustness score which allows us to properly compare OODD approaches. Then, we expand this formulation to a robustness score that allows finding the best optimization method for a given ID dataset and OODD approach.

Let $X$ be the random variable representing the value of an OODD metric (e.g., AUROC) whose distribution is represented by a probability density function (PDF) $p(x)$. An OODD metric depends on: 
\begin{enumerate}
    \item a given OODD approach $m$ (e.g., max-softmax),
    \item an ID dataset $i$ (e.g., MNIST),
    \item an OOD dataset $o$ (e.g., Fashion-MNIST).
\end{enumerate} 
Thus, we consider its conditional distribution with the PDF $p(x| i, o, m)$ rather than $p(x)$. For conciseness, we denote $\zeta := (i,o,m)$. Then, $p(x| \zeta) \equiv p(x| i, o, m)$ can be represented by the following mixture PDFs \citep{Yaakov_mixture_pdf}: 
\begin{equation}
    p(x|\zeta) = \sum_{t \in \mathcal{T}} w^{\zeta,t} \cdot p(x | \zeta, t)
\end{equation}
where $\mathcal{T}$ represents a set of optimization methods that
are used to train ID models in this paper; $p(x | \zeta, t)$ is the conditional PDF of the OODD metric when the optimization method $t \in \mathcal{T}$ is used to train the ID model; $w^{\zeta,t} \geq 0$ is the weight of $p(x | \zeta, t)$, summing up to unity (i.e., $\sum_{t \in \mathcal{T}}w^{\zeta,t}=1$). Please note that we considered a PDF for a given $\zeta$ and $t$ because there is no guarantee that the ID model with the given optimization method will converge to the same solution all the time. This is due to factors such as randomness in initial weights and data in dataset.

$p(x | \zeta, t)$ is unknown, but its empirical mean and variance can be approximated. This is achieved by training the same ID model with the same optimization method $t$ but different random initial weights. Thus, the mean and variance of $X$ conditioned by $\zeta$ and an optimization method $t$ are empirically approximated by:
\begin{align}
\label{mean_opt_o}
\mathbb{E}[X|\zeta,t] &\approx \mu^{\zeta,t} := \frac{1}{N}\sum_{j=1}^{N} X^{\zeta,t}_j
\\
\label{var_opt_o}
\mathrm{Var}[X|\zeta,t] &\approx \mathrm{Var}^{\zeta,t} := \frac{1}{N}\sum_{j=1}^{N} (\, X^{\zeta,t}_j - \mu^{\zeta,t} \,)^2
\end{align}
in which $N$ is the number of the ID models trained with the ID dataset $i$ and optimization method $t$ but $N$ different random initial weights. $X^{\zeta,t}_j$ represents the OODD metric  calculated for the $j$\textit{-th} ID model of the optimization method $t$.

In order to obtain $w^{\zeta,t}$, we first define confidence score $c^{\zeta,t}$ as:

\begin{equation}
c^{\zeta,t} := \frac{1}{\sqrt{ \mathrm{Var}^{\zeta,t} }+ \epsilon}
\end{equation}
The confidence score $c^{\zeta,t}$ specifies how much the OODD metric calculated for the ID model trained by the optimization method $t$ can be trusted. $\epsilon > 0$ is a sufficiently small value used for numerical stability when $\mathrm{Var}^{\zeta,t}=0$. Then, the normalized form of the confidence score is used to calculate $w^{\zeta,t}$:
\begin{equation}
\label{weight_t}
w^{\zeta,t} = \frac{c^{\zeta,t}}{ \sum_{t \in \mathcal{T}} c^{\zeta,t} } 
\end{equation}
Having the empirical mean $\mu^{\zeta, t}$ and variance $\mathrm{Var}^{\zeta, t}$ of the OODD metric $X$ conditioned by $\zeta$ and $t$, we obtain
those of $X$ conditioned only by $\zeta$ using the weights $w^{\zeta, t}$ as \citep{Yaakov_mixture_pdf}:
\begin{align}
\label{mean_o_i}
\mathbb{E}[X|\zeta] &\approx \mu^{\zeta} := \sum_{t \in \mathcal{T}} w^{\zeta,t} \cdot \mu^{\zeta,t}
\\
\notag
\mathrm{Var}[X|\zeta] &\approx \mathrm{Var}^{\zeta} 
\\
\label{var_o_i}
&:= \sum_{t \in \mathcal{T}} w^{\zeta,t} \cdot \big(\mathrm{Var}^{\zeta, t} + (\mu^{\zeta} - \mu^{\zeta,t})^2\big)
\end{align}
The approximated $\mu^{\zeta}$ and $\mathrm{Var}^{\zeta}$ represent how good the corresponding OODD approach according to the OODD metric $X$ on average and how much variation is expected to be seen in the value of $X$, respectively. However, it is still challenging to compare OODD approaches because an OODD approach might have a high mean and high variance while another one might have a slightly less mean but very low variance. Although the former one has a high mean (i.e., better detection), its high variation means that it might show worse results for some optimization methods. The later one gives highly consistent detection performance, but it does not have the best mean.

To accommodate these cases (i.e., the trade-off between mean and variance)  and be able to quantitatively compare OODD approaches, we propose using the coefficient of variation (CV) (conditioned by $\zeta$) as a final robustness score, defined as:
 \begin{equation}
 \label{cv_o_i}
CV^{\zeta} := \frac{\sqrt{\mathrm{Var}^{\zeta} }} { \mu^{\zeta} }
\end{equation}
The $CV$ takes into account both mean and variance. It favors an OODD approach that has a high mean and low variance for an OODD metric. Therefore, a lower $CV^{\zeta}$ value corresponds to lower variability in the mean of the OODD metric. A robust OODD approach is expected to have a low $CV^{\zeta}$ value. $CV^{\zeta}$ can be calculated for any OODD metrics as we demonstrate it in the next section.

In Figure \ref{fashion-mnist_md_max-softmax_optimizers_comp}, we demonstrated that an OODD approach can better detect OOD samples if the ID model has been trained with a specific optimization method. $\mu^{\zeta, t}$ and $\mathrm{Var}^{\zeta,t}$ provide a way to find the best optimization method for a given ID, OOD, and OODD approach. We can extend our proposal to finding the best optimization method for a given OODD approach and ID dataset. For this purpose, we consider conditional PDF $p(x|i,m,t)$. For conciseness, we denote $ \xi:= (i,m,t)$. Then:
\begin{equation}
p(x|\xi) = \sum_{o \in \mathcal{O}_i} w^{\xi,o} \cdot p(x| \xi, o)
\end{equation}

$\mathcal{O}_i$ is a set of OOD datasets available for the ID dataset $i$. This constraint emerges from not being able to represent all OOD inputs for an ID dataset and also how OODD approaches have been compared in literature. Since $ p(x| \xi, o) \equiv  p(x | \zeta, t)$ then the mean and variance of $X$ conditioned by $\xi$ and $o$ can be approximated based on \eqref{mean_opt_o} and \eqref{var_opt_o}. Similarly, $w^{\xi,o} \equiv w^{\zeta,t}$ so $w^{\xi,o}$ can be calculated based on \eqref{weight_t}. The mean and variance of $X$ conditioned by $\xi$ can then be approximated in a similar way to \eqref{mean_o_i} and \eqref{var_o_i}, but their summation is over $\mathcal{O}_i$ instead of $ \mathcal{T}$. $CV^{\xi}$, which represents robustness score for optimization method $t$, can now be used to select the best optimization method for a given OODD approach and ID dataset.

\section{EXPERIMENTS}
We compare several OODD approaches according to our proposed robustness score. Furthermore, we find optimization methods that give better detection performance for a given ID dataset and OODD approach.

\subsection{ID AND OOD DATASETS}

We use MNIST and Fashion-MNIST (or F-MNIST) as our ID datasets. MNIST is a dataset of  $28\times28$ grayscale images of handwritten digits. It has $60,000$ training images and $10,000$ test images. Fashion-MNIST is similar to the MNIST dataset by also containing $28\times28$ grayscale images, but the images represent Zalando's articles \citep{FMNIST_2017}. We used a convolutional neural network (CNN) with two convolutional layers and two fully connected layers and trained one model for each dataset. The CNN includes dropout layers that allow us to use uncertainty-estimate techniques for the detection of OOD inputs as well.

The OOD datasets of MNIST ($\mathcal{O}_{MNIST}$) are  Fashion-MNIST, Omniglot, Gaussian noise, and Uniform noise. The OOD datasets of Fashion-MNIST ($\mathcal{O}_{F-MNIST}$) are MNIST, Omniglot, Gaussian noise, and Uniform noise. We use test datasets of MNIST, Fashion-MNIST, and Omniglot for calculating OODD metrics. We keep the number of ID and OOD inputs the same during evaluation (by randomly selecting data from the larger dataset to match the number of instances in the smaller dataset). The details of the aforementioned OOD datasets are as follows:
\begin{itemize}
     \item \textbf{Omniglot} contains different handwritten characters from 50 different alphabets. The images have been downsampled to $28 \times28$ images \citep{omniglot_Lake1332}. 
     \item \textbf{Gaussian noise} includes random normal noise with $\mu = 0.5$ and $\sigma = 1$, clipped to [0, 1].
     \item \textbf{Uniform noise} includes random uniform noise between [0, 1].
\end{itemize}

\subsection{OPTIMIZATION METHODS AND TRAINING PROCESS} \label{opt_methods}
We used seven optimization methods for our experiments. These methods are Adam \citep{Adam_Kingma_2015}, RMSprop \citep{Tieleman_RmsProp_2012}, Adamax  \citep{Adam_Kingma_2015}, Nadam \citep{Dozat_nadam_2016}, SGD, Adagrad \citep{Adagrad_Duchi_2011}, and Adadelta \citep{Zeiler_adadelta_2012}. The values of parameters used for these optimization methods are depicted in the Appendix, Table \ref{opt_name_values_para}. They are values recommended by the tools we used for our experiments, which are Tensorflow and Keras. 

To train models, we used the early stop technique in which the training process is stopped once no improvements are seen in the validation loss (the chosen patience parameter is 10 epochs). However, the saved model after the training process stops is the model that has the best validation loss. This has been chosen to avoid overfitting.

\subsection{OODD METRICS}\label{evaluation-metrics}
There are different metrics to evaluate the performance of OODD approaches. We adopted the following metrics \citep{Hendrycks-baseline-2017,Liang-ODIN-2018}.
\begin{itemize}
 \item FPR at $95\%$ TPR is the probability of an out-of-distribution (i.e., negative) input being misclassified as in-distribution (i.e., positive) input when the true positive rate (TPR) is as high as $95\%$. True positive rate is calculated by $\textit{TPR} = \textit{TP} / (\textit{TP}+\textit{FN})$, where TP and FN denote true positives and false negatives, respectively. The false positive rate (FPR) is computed by $\textit{FPR} = \textit{FP} / (\textit{FP}+\textit{TN})$, where FP and TN denote false positives and true negatives, respectively.
 \item Detection error calculates the misclassification probability when TPR is $95\%$. It is equal to $0.5*(1 - \textit{TPR}) + 0.5*\textit{FPR}$, where we assume that both positive and negative examples have an equal probability of appearing in the evaluation test.
 \item AUROC is the Area Under the Receiver Operating Characteristic curve. It is interpreted as the probability that a positive example is assigned a higher detection score than a negative example. An ideal OOD detector expects an AUROC score of $100\%$.
 \item AUPR is the Area under the Precision-Recall curve. It is a graph reflecting precision equal to $\textit{TP}/(\textit{TP}+\textit{FP})$ and recall equal to $\textit{TP}/(\textit{TP}+\textit{FN})$ against each other. The metric AUPR-In and AUPR-Out represent the area under the precision-recall curve where in-distribution or out-of-distribution images are specified as positives, respectively.
\end{itemize}

\begin{table*}
    \centering
    \caption{$\mu^{\zeta,t}$ and $\mathrm{Var}^{\zeta,t}$ when $\zeta=($MNIST, Fashion-MNIST, max-softmax$)$ and $t = $Adam. $\downarrow$ indicates that lower values are better; $\uparrow$ indicates that higher values are better.}
    \begin{tabular}{|l|l|l|l|l|l|}
    \hline
    
    \multicolumn{1}{|c|}{\begin{tabular}[c]{@{}c@{}}Opt. \\method \end{tabular}} &  
    \multicolumn{1}{|c|}{\begin{tabular}[c]{@{}c@{}}FPR at\\ 95\% TPR $\downarrow$ \end{tabular}}&
    \multicolumn{1}{c|}{\begin{tabular}[c]{@{}c@{}}Detection \\ error $\downarrow$    \end{tabular}} & 
    \multicolumn{1}{|c|}{\begin{tabular}[c]{@{}c@{}}AUROC \\ $\uparrow$ \end{tabular}}&
    \multicolumn{1}{c|}{\begin{tabular}[c]{@{}c@{}}AUPR\\Out \\ $\uparrow$ \end{tabular}} & 
    \multicolumn{1}{|c|}{\begin{tabular}[c]{@{}c@{}}AUPR\\In \\ $\uparrow$ \end{tabular}} \\ \hline
  Adam & 10.24 & 7.61 & 97.728 & 97.981 & 97.483 \\
  Adam & 7.47 & 6.225 & 97.834 & 98.272 & 97.176 \\
  Adam & 16.5 & 10.735 & 96.421 & 96.705 & 96.018 \\
  Adam & 7.57 & 6.18 & 98.195 & 98.482 & 97.897 \\
  Adam & 15.32 & 10.11 & 96.554 & 96.671 & 96.277 \\
\hline
 $\mu^{\zeta,t} | \mathrm{Var}^{\zeta,t}$  & 11.42 | 14.567 & 8.172 | 3.68 & 97.346 | 0.518 & 97.622  | 0.607 & 96.97 | 0.51 \\ \hline
\end{tabular}
\label{mnist_fmnist_r_1}
\end{table*}
\begin{table*}
    \centering
    \caption{$\mu^{\zeta}$ and $\mathrm{Var}^{\zeta}$ when $\zeta=($MNIST, Fashion-MNIST, max-softmax$)$ and $\mathcal{T} =\{$Adam, RMSprop, Adamax, Nadam, SGD, Adagrad, Adadelta$\}$. $\downarrow$ indicates that lower values are better; $\uparrow$ indicates that higher values are better.}
    \begin{tabular}{|l|l|l|l|l|l|}
    \hline

    \multicolumn{1}{|c|}{\begin{tabular}[c]{@{}c@{}}Opt. \\ Method \end{tabular}} &  
    \multicolumn{1}{|c|}{\begin{tabular}[c]{@{}c@{}}FPR at\\ 95\% TPR \\ $\mu \downarrow | \mathrm{Var} \downarrow$ \end{tabular}}&
    \multicolumn{1}{c|}{\begin{tabular}[c]{@{}c@{}}Detection \\ error \\ $\mu \downarrow | \mathrm{Var}\downarrow$    \end{tabular}} & 
    \multicolumn{1}{|c|}{\begin{tabular}[c]{@{}c@{}}AUROC \\ $\mu \uparrow | \mathrm{Var} \downarrow$ \end{tabular}}&
    \multicolumn{1}{c|}{\begin{tabular}[c]{@{}c@{}}AUPR\\Out \\ $\mu\uparrow | \mathrm{Var} \downarrow$ \end{tabular}} & 
    \multicolumn{1}{|c|}{\begin{tabular}[c]{@{}c@{}}AUPR\\In \\ $\mu\uparrow | \mathrm{Var} \downarrow$ \end{tabular}} \\ \hline
 Adam & 11.42 | 14.567 & 8.172 | 3.68 & 97.346 | 0.518 & 97.622 | 0.607 & 96.97 | 0.51 \\
 RMSprop & 7.804 | 5.236 & 6.319 | 1.446 & 97.988 | 0.256 & 98.278 | 0.276 & 97.609 | 0.344 \\
 Adamax & 8.844 | 4.348 & 6.897 | 1.096 & 97.608 | 0.156 & 97.971 | 0.129 & 97.111 | 0.224 \\
 Nadam & 9.018 | 0.886 & 6.968 | 0.231 & 97.751 | 0.022 & 98.054 | 0.026 & 97.349 | 0.118 \\
 SGD & 7.236 | 3.911 & 6.042 | 1.039 & 98.026 | 0.204 & 98.385 | 0.182 & 97.56 | 0.212 \\
 Adagrad & 8.56 | 3.697 & 6.721 | 0.972 & 97.685 | 0.233 & 98.103 | 0.176 & 97.122 | 0.351 \\
 Adadelta & 8.252 | 9.935 & 6.572 | 2.634 & 97.88 | 0.503 & 98.191 | 0.455 & 97.48 | 0.664 \\
\hline
 $\mu^{\zeta} | \mathrm{Var}^{\zeta}$ &8.634 | 5.506 &6.769 | 1.445 &97.756 | 0.219 &98.089 | 0.216 &97.315 | 0.349  \\\hline
\end{tabular}
\label{mnist_fmnist_r_2}
\end{table*}
\begin{table*}
    \centering
    \caption{$\mu^{\zeta}$ and $\mathrm{Var}^{\zeta}$ when $\zeta=($MNIST, Fashion-MNIST, $m )$ and $m \in \{$max-softmax, ODIN,MD, Entropy, Margin, MC-D, MI $\}$. $\downarrow$ indicates that lower values are better; $\uparrow$ indicates that higher values are better.}
    \begin{tabular}{|l|l|l|l|l|l|}
   \hline
    \multicolumn{1}{|c|}{\begin{tabular}[c]{@{}c@{}}OODD \\ approaches \end{tabular}} &  
    \multicolumn{1}{|c|}{\begin{tabular}[c]{@{}c@{}}FPR at\\ 95\% TPR \\ $\mu \downarrow | \mathrm{Var} \downarrow$ \end{tabular}}&
    \multicolumn{1}{c|}{\begin{tabular}[c]{@{}c@{}}Detection \\ error \\ $\mu \downarrow | \mathrm{Var}\downarrow$    \end{tabular}} & 
    \multicolumn{1}{|c|}{\begin{tabular}[c]{@{}c@{}}AUROC \\ $\mu \uparrow | \mathrm{Var} \downarrow$ \end{tabular}}&
    \multicolumn{1}{c|}{\begin{tabular}[c]{@{}c@{}}AUPR\\Out \\ $\mu\uparrow | \mathrm{Var} \downarrow$ \end{tabular}} & 
    \multicolumn{1}{|c|}{\begin{tabular}[c]{@{}c@{}}AUPR\\In \\ $\mu\uparrow | \mathrm{Var} \downarrow$ \end{tabular}} \\ \hline
 max-softmax &8.634 | 5.506 &6.769 | 1.445 &97.756 | 0.219 &98.089 | 0.216 &97.315 | 0.349 \\
 ODIN &4.932 | 3.081 &4.84 | 0.983 &98.944 | 0.12 &99.036 | 0.104 &98.87 | 0.137 \\
 MD &64.707 | 31.994 &34.837 | 7.966 &69.108 | 19.587 &75.163 | 14.237 &62.276 | 18.813 \\
 Entropy &8.549 | 5.467 &6.728 | 1.442 &97.944 | 0.22 &98.195 | 0.202 &97.703 | 0.327 \\
 Margin &8.777 | 5.52 &6.842 | 1.448 &97.625 | 0.214 &98.023 | 0.222 &96.855 | 0.344 \\
 MC-D &8.218 | 4.33 &6.536 | 1.209 &97.868 | 0.221 &98.213 | 0.191 &97.465 | 0.298 \\
 MI &8.817 | 4.748 &6.812 | 1.285 &97.238 | 0.224 &97.857 | 0.199 &96.125 | 0.458  \\\hline
\end{tabular}
\label{mnist_fmnist_r_3}
\end{table*}
\begin{table*}
    \centering
    \caption{The $CV^{\zeta}$ values for the $\mu^{\zeta}$ and $\mathrm{Var}^{\zeta}$ in Table \ref{mnist_fmnist_r_3}. Since better FPR@95\%TPR and detection error should have low mean, we use the inverse of $\mathrm{Var}^{\zeta}$ to calculate $CV^{\zeta}$  to have consistent score among all other metrics. The bold values highlights the most robust OODD approach for the given metric. $\downarrow$ indicates that lower values are better.}
    \begin{tabular}{|l|l|l|l|l|l|}
    \hline
    \multicolumn{1}{|c|}{\begin{tabular}[c]{@{}c@{}}OODD \\approaches \end{tabular}} &  
    \multicolumn{1}{|c|}{\begin{tabular}[c]{@{}c@{}}FPR at\\ 95\% TPR \\ $\mu * \sqrt{\mathrm{Var}} \downarrow $ \end{tabular}}&
    \multicolumn{1}{c|}{\begin{tabular}[c]{@{}c@{}}Detection \\ error \\ $\mu * \sqrt{\mathrm{Var}} \downarrow $ \end{tabular}} & 
    \multicolumn{1}{|c|}{\begin{tabular}[c]{@{}c@{}}AUROC\\ $\sqrt{\mathrm{Var}} / \mu \downarrow $ \end{tabular}}&
    \multicolumn{1}{c|}{\begin{tabular}[c]{@{}c@{}}AUPR\\Out$\sqrt{\mathrm{Var}} / \mu \downarrow $ \end{tabular}} & 
    \multicolumn{1}{|c|}{\begin{tabular}[c]{@{}c@{}}AUPR\\In $\sqrt{\mathrm{Var}} / \mu \downarrow $ \end{tabular}} \\ \hline
 max-softmax &20.258 &8.138 &0.005 &0.005 &0.006  \\
 \textbf{ODIN} & \textbf{ 8.657 }  & \textbf{ 4.797 }  & \textbf{ 0.003 }  & \textbf{ 0.003 }  & \textbf{ 0.004 }  \\ 
 MD &365.988 &98.313 &0.064 &0.05 &0.07 \\
 Entropy &19.99 &8.082 &0.005 &0.005 &0.006 \\
 Margin &20.62 &8.232 &0.005 &0.005 &0.006 \\
 MC-D &17.104 &7.191 &0.005 &0.004 &0.006 \\
 MI &19.214 &7.726 &0.005 &0.005 &0.007  \\\hline
\end{tabular}
\label{mnist_fmnist_r_4}
\end{table*}
\begin{table}
\centering
\caption{Best OODD approaches based on FPR@95\%TPR for different ID and OOD datasets.}
    \begin{tabular}{lcc}
        \toprule
        \multicolumn{1}{c}{\begin{tabular}[c]{@{}c@{}}ID\end{tabular}} &
        \multicolumn{1}{c}{\begin{tabular}[c]{@{}c@{}}OOD\end{tabular}} &
        \multicolumn{1}{c}{\begin{tabular}[c]{@{}c@{}}OODD\\ approach\end{tabular}}\\
        \bottomrule
        
                & F-MNIST & ODIN\\
        MNIST   & Omniglot & MC-D\\
                & Gaussian & ODIN\\
                & Uniform & ODIN\\ 
               \hline
                & MNIST & MD\\
        F-MNIST & Omniglot &MI \\
                & Gaussian & ODIN\\
                & Uniform & MI\\ 
        \bottomrule
    \end{tabular}
    \label{top_ood_detection}
\end{table}

\subsection{OODD APPROACHES} \label{ood_approaches}
In this paper, we focus on OODD approaches that apply to already-trained models and are easiest to incorporate into existing application. There are other OODD approaches such as generative or classifier-based approaches \citep{aaeOOD,Wang_safe_classification2017, Kimin_Lee-2018} that require models to be retrained. We consider these as our future work. The OODD approaches used in our experiments are max-softmax \citep{Hendrycks-baseline-2017}, ODIN (without input preprocessing and $T$=1000) \citep{Liang-ODIN-2018}, Mahalanobis Distance (MD) based on logits features (without input preprocessing) \citep{Lee-mah-2019}, and Entropy, Margin, the MC-dropout (MC-D) \citep{Gal-dropout-2016}, and Mutual Information (MI) \citep{Gal_bald_2017}.

\subsection{ROBUSTNESS SCORES OF OODD APPROACHES}
Table ~\ref{mnist_fmnist_r_1} shows the means $\mu^{\zeta,t}$ and variances $\mathrm{Var}^{\zeta,t}$ of different OODD metrics (the last row) for MNIST as ID, Fashion-MNIST as OOD, max-softmax as the OODD approach, and Adam as the optimization method. For example, the mean value for FPR@95\%TPR is 11.42 and we might have better or worse FPR@95\%TPR under a certain solution obtained by Adam because of the high variance, which is 14.567. To approximate the means and variances based on \eqref{mean_opt_o} and \eqref{var_opt_o}, OODD metrics are calculated for five models (each represented by a row in the middle cell) with different initial weights. FPR@95\%TPR and detection error show the highest variance among other metrics. High variance for other OODD metrics can also be seen in our other experiment in Table ~\ref{fmnist_mnist_r_1} in the Appendix.

Table ~\ref{mnist_fmnist_r_2} depicts the means $\mu^{\zeta}$ and variances $\mathrm{Var}^{\zeta}$ of different OODD metrics (the last row) when $\zeta=($MNIST, Fashion-MNIST, max-softmax$)$. For example, the mean value for FPR@95\%TPR is 8.63 and we expect to see some difference in the performance due to its variance, which is 5.506. The means and variances are regardless of which optimization method is selected to train the ID model. To approximate these moments (based on \eqref{mean_o_i} and \eqref{var_o_i}), we use the mean and variance of the metrics calculated for seven optimization methods (each represented by a row in the middle cell) in the same way as depicted in Table ~\ref{mnist_fmnist_r_1}. We can observe that SGD has the lowest mean for FPR@95\%TPR, but it does not have the lowest variance. Nadam has the lowest variance but not the highest mean. Adam has the highest mean, but also the highest variance.

The mean and variance of OODD metrics for different OODD approaches are depicted in Table \ref{mnist_fmnist_r_3}. MD shows a poor detection performance in comparison to others, but in our other experiments in the Appendix (Table ~\ref{fmnist_mnist_r_3}), MD shows good performance. This can be related to the fact that the distribution of features for the logits layer in the MNIST model may not follow a multivariate Gaussian distribution. Therefore, the OOD score comes our of MD is not reliable for separating ID and OOD inputs. The MD approach may also be an ID dataset dependent.

To find the best OODD approach based on their mean and variance, the  $CV^{\zeta}$ values of different metrics and OODD approaches from Table \ref{mnist_fmnist_r_3} are shown in Table ~\ref{mnist_fmnist_r_4}. ODIN shows the lowest  $CV^{\zeta}$ scores for all OODD metrics and can be considered as a robust OODD approach for MNIST as ID and Fashion-MNIST as OOD. A summary of the best OODD approach for other ID and OO datasets based on FPR@95\%TPR is depicted in Table \ref{top_ood_detection}.

We can observe that ODIN performs better than others on average; however, there is not a perfect OODD approach. This is opposite to what we can see in many published papers where authors compare their proposed OODD approach with others and theirs outperforms all others. This demonstrates how our robustness score could incorporate the effect of optimization methods in OODD metrics for a sound comparison. 

\subsection{ROBUST OPTIMIZATION METHODS}
Table \ref{mnist_fmnist_o_1} show the mean $\mu^{\xi}$ and variance of $\mathrm{Var}^{\xi}$ of different OODD metrics (the last row) when $\xi=($MNIST, max-softmax, Adam$)$. These values show the quality of solutions offered by Adam for detecting different kinds of OOD datasets for a given ID and OODD approach. The moments are approximated over metrics calculated for different OOD datasets. These metrics (rows in the middle cell) are approximated in similar way as in Table \ref{mnist_fmnist_r_1}. Table \ref{mnist_fmnist_o_2} shows $\mu^{\xi}$ and $\mathrm{Var}^{\xi}$ when $\xi=($MNIST, max-softmax, $ t)$ and $t$ can be one of the seven optimization methods. To find the best optimization method for a given ID and OODD approach based on the means and variances, the $CV^{\xi}$ values are shown in Table \ref{mnist_fmnist_o_3}. SGD and Nadam have the lowest scores for all metrics. It means the MNIST models trained with these optimization methods can provide better solutions for detecting OOD inputs when max-softmax is used as the OODD approach.
Table \ref{top_two_opt_methods} shows a summary of top-two optimization methods based on FPR@95\%TPR for two ID datasets and several OODD approaches.

\begin{table*}
    \centering
    \caption{$\mu^{\xi}$ and $\mathrm{Var}^{\xi}$ when $\xi=($MNIST, max-softmax, Adam$)$ and $\mathcal{O}_{MNIST} = \{$Fashion-MNIST, Omniglot, Gaussian, Uniform$\}$ }
    \begin{tabular}{|l|l|l|l|l|l|}
    \hline
    \multicolumn{1}{|c|}{\begin{tabular}[c]{@{}c@{}}OOD\end{tabular}} &  
    \multicolumn{1}{|c|}{\begin{tabular}[c]{@{}c@{}}FPR at\\ 95\% TPR \\ $\mu \downarrow | \mathrm{Var} \downarrow$ \end{tabular}}&
    \multicolumn{1}{c|}{\begin{tabular}[c]{@{}c@{}}Detection \\ error \\ $\mu \downarrow | \mathrm{Var}\downarrow$    \end{tabular}} & 
    \multicolumn{1}{|c|}{\begin{tabular}[c]{@{}c@{}}AUROC \\ $\mu \uparrow | \mathrm{Var} \downarrow$ \end{tabular}}&
    \multicolumn{1}{c|}{\begin{tabular}[c]{@{}c@{}}AUPR\\Out \\ $\mu\uparrow | \mathrm{Var} \downarrow$ \end{tabular}} & 
    \multicolumn{1}{|c|}{\begin{tabular}[c]{@{}c@{}}AUPR\\In \\ $\mu\uparrow | \mathrm{Var} \downarrow$ \end{tabular}} \\ \hline
     F-MNIST & 11.42 | 14.567 & 8.172 | 3.68 & 97.346 | 0.518 & 97.622 | 0.607 & 96.97 | 0.51 \\
     Omniglot & 6.08 | 1.373 & 5.246 | 0.547 & 98.205 | 0.127 & 98.613 | 0.079 & 97.559 | 0.319 \\
     Gaussian & 1.146 | 1.067 & 0.99 | 0.343 & 99.348 | 0.543 & 99.62 | 0.167 & 98.073 | 5.747 \\
     Uniform  & 3.368 | 1.663 & 2.757 | 0.854 & 98.406 | 0.567 & 99.003 | 0.19 & 96.415 | 5.02 \\
\hline
$\mu^{\xi} | \mathrm{Var}^{\xi}$ &4.162 | 11.733 &3.437 | 6.649 &98.282 | 0.78 &98.82 | 0.579 &97.257 | 1.683  \\\hline
\end{tabular}
\label{mnist_fmnist_o_1}
\end{table*}
\begin{table*}
    \centering
    \caption{$\mu^{\xi}$ and $\mathrm{Var}^{\xi}$ when $\xi$ = (MNIST,max-softmax, $t$), $t \in \mathcal{T} =\{$Adam, RMSprop, Adamax, Nadam, SGD, Adagrad, Adadelta$\}$ and $\mathcal{O}_{MNIST} = \{$Fashion-MNIST, Omniglot, Gaussian, Uniform$\}$  }
    \begin{tabular}{|l|l|l|l|l|l|}
    \hline

    \multicolumn{1}{|c|}{\begin{tabular}[c]{@{}c@{}}Opt. \\method \end{tabular}} &  
    \multicolumn{1}{|c|}{\begin{tabular}[c]{@{}c@{}}FPR at\\ 95\% TPR \\ $\mu \downarrow | \mathrm{Var} \downarrow$ \end{tabular}}&
    \multicolumn{1}{c|}{\begin{tabular}[c]{@{}c@{}}Detection \\ error \\ $\mu \downarrow | \mathrm{Var}\downarrow$    \end{tabular}} & 
    \multicolumn{1}{|c|}{\begin{tabular}[c]{@{}c@{}}AUROC \\ $\mu \uparrow | \mathrm{Var} \downarrow$ \end{tabular}}&
    \multicolumn{1}{c|}{\begin{tabular}[c]{@{}c@{}}AUPR\\Out \\ $\mu\uparrow | \mathrm{Var} \downarrow$ \end{tabular}} & 
    \multicolumn{1}{|c|}{\begin{tabular}[c]{@{}c@{}}AUPR\\In \\ $\mu\uparrow | \mathrm{Var} \downarrow$ \end{tabular}} \\ \hline
 Adam &4.162 | 11.733 &3.437 | 6.649 &98.282 | 0.78 &98.82 | 0.579 &97.257 | 1.683 \\
 RMSprop &4.059 | 18.879 &3.497 | 9.398 &98.443 | 1.862 &98.895 | 1.189 &97.696 | 4.433 \\
 Adamax &3.487 | 10.255 &3.305 | 6.656 &98.674 | 0.88 &99.058 | 0.609 &98.014 | 1.364 \\
 Nadam &2.404 | 9.64 &2.495 | 6.995 &99.225 | 0.86 &99.476 | 0.518 &99.293 | 1.018 \\
 SGD &2.82 | 6.985 &2.578 | 5.025 &98.849 | 0.632 &99.22 | 0.404 &98.161 | 1.188 \\
 Adagrad &4.346 | 9.209 &3.835 | 5.834 &98.417 | 0.816 &98.843 | 0.502 &97.702 | 2.285 \\
 Adadelta &4.188 | 12.394 &3.707 | 6.758 &98.406 | 1.241 &98.863 | 0.716 &97.554 | 3.719  \\\hline
\end{tabular}
\label{mnist_fmnist_o_2}
\end{table*}

\begin{table*}
    \centering
    \caption{ The $CV^{\xi}$ values for the $\mu^{\xi}$ and $\mathrm{Var}^{\xi}$ efined in Table \ref{mnist_fmnist_o_2}. Since better FPR@95\%TPR and detection error should have low mean, we use the inverse of $\mathrm{Var}^{\xi}$ to calculate $CV^{\xi}$  to have consistent score among all other metrics. The bold values highlights the most robust optimization method for the given metric. $\downarrow$ indicates that lower values are better.  }
    \begin{tabular}{|l|l|l|l|l|l|}
    \hline

    \multicolumn{1}{|c|}{\begin{tabular}[c]{@{}c@{}}Opt. \\ methods \end{tabular}} &  
    \multicolumn{1}{|c|}{\begin{tabular}[c]{@{}c@{}}FPR at\\ 95\% TPR \\ $\mu * \sqrt{\mathrm{Var}} \downarrow $ \end{tabular}}&
    \multicolumn{1}{c|}{\begin{tabular}[c]{@{}c@{}}Detection \\ error \\ $\mu * \sqrt{\mathrm{Var}} \downarrow $ \end{tabular}} & 
    \multicolumn{1}{|c|}{\begin{tabular}[c]{@{}c@{}}AUROC  \\ $\sqrt{\mathrm{Var}} / \mu \downarrow $ \end{tabular}}&
    \multicolumn{1}{c|}{\begin{tabular}[c]{@{}c@{}}AUPR\\Out \\ $\sqrt{\mathrm{Var}} / \mu \downarrow $ \end{tabular}} & 
    \multicolumn{1}{|c|}{\begin{tabular}[c]{@{}c@{}}AUPR\\In \\ $\sqrt{\mathrm{Var}} / \mu \downarrow $ \end{tabular}} \\ \hline
 Adam &14.258 &8.867 &0.009 &0.008 &0.013 \\ 
 RMSprop &17.641 &10.725 &0.014 &0.011 &0.022  \\ 
 Adamax &11.169 &8.529 &0.01 &0.008 &0.012  \\
 Nadam &7.468 &6.602 &0.009 &0.007 & \textbf{ 0.01 }   \\
 \textbf{SGD} & \textbf{ 7.456 }  & \textbf{ 5.782 }  & \textbf{ 0.008 }  & \textbf{ 0.006 }  & 0.011  \\
 Adagrad &13.193 &9.264 &0.009 &0.007 &0.015  \\
 Adadelta &14.749 &9.641 &0.011 &0.009 &0.02  \\\hline
\end{tabular}
\label{mnist_fmnist_o_3}
\end{table*}


\begin{table}
\centering
\caption{Top-two optimization methods based on FPR@95\%TPR for OODD approaches.}
    \begin{tabular}{lcc}
        \toprule
        \multicolumn{1}{c}{\begin{tabular}[c]{@{}c@{}}OODD\\ approach\end{tabular}} &
        \multicolumn{1}{c}{\begin{tabular}[c]{@{}c@{}}MNIST\end{tabular}} &
        \multicolumn{1}{c}{\begin{tabular}[c]{@{}c@{}}Fashion-MNIST\end{tabular}}\\
        \bottomrule
        
        max-softmax   & SGD,Nadam & RMSProp,SGD\\
        ODIN          & Nadam,Adamax &RMSProp,Nadam \\
        MD            & MSprop,Adagrad & Nadam,Adagrad\\
        Entropy       & Nadam,SGD & RMSProp,Nadam\\ 
        Margin        & Nadam,SGD & RMSProp,SGD\\
        MC-D          & Nadam,SGD & RMSProp,Adamax\\
        MI            & Adagrad,Nadam& Adamax,Adadelta\\
        
        \bottomrule
    \end{tabular}
    
    \label{top_two_opt_methods}
\end{table}

\subsection{DISCUSSION}
The robustness of OODD approaches under optimization methods is a significant aspect that needs to be incorporated into OOD detection metrics. We proposed a robustness score that takes into account the effect of optimization methods. It gives equal importance to both mean and variance. However, we can imagine in some context and application it might be preferred to pay more attention to either mean or variance. In that case, instead of using the $CV$ \eqref{cv_o_i}, a custom formula can be adopted. 

We calculated our robustness score for five OODD metrics and this approach can be easily adapted for any custom OODD metric. We specifically focused on FPR@95\%TPR to describe the application of our robustness score and our conclusions are based on this metric. Focusing on other OODD metrics can change some of our conclusions (e.g., the best optimization method for an OODD approach). Our goal in this paper is not to extensively compare OODD approaches or recommend one. The goal is to demonstrate the importance of optimization methods in the domain of OODD and suggest a way to incorporate it into OODD metrics.

We observed that some OODD approaches better detect OOD inputs if the ID model has been trained with a specific optimization method. We experimentally found the best-performing optimization methods for several OODD approaches based on FPR@95\%TPR. However, it is still unclear what solutions these optimization methods provide that cause these approaches to work better.

One hypothesis is that these optimization methods achieve solutions that generalize better. Thus, OODD approaches are able to better detect OOD and ID inputs. \cite{Wilson_Ashia_2017} studied the effect of adaptive optimization methods and SGD on the generalization of solutions. They found that solutions offered by SGD generalize better. SGD is among the top optimization methods found for OODD approaches in this paper. However, it is not the best for all ID and OODD approaches. We believe this aspect of optimization methods requires rigorous studies.

The second hypothesis is that an OODD approach has specific requirements on the ID model to demonstrate its best detection performance. For example, the MD approach requires the class-conditional distribution of features to be Gaussian. This condition might happen under specific optimization methods. Max-softmax and uncertainty-based approaches need confident models and might happen under specific optimization methods, too. Therefore, it is important for OODD approaches to precisely define the conditions under which they achieve their claimed performance and provide appropriate tests for practitioners to apply before using the approach. 

Please note that one could evaluate pairs, i.e., a detection appraoch and optimizer; by doing this, the winning approach is changed for every ID and OOD data pair. Therefore, it becomes challenging to compare OODD approaches because there is no winner. Our weighted average also applies to pairwise evaluation (to average out the effects over different ID and OOD data combinations) and can show the overall robust OODD approach with a statistically rigorous formulation. Moreover, when comparing OODD approaches applied to an already-trained ID model, the ID model should be kept the same for a fair comparison. Thus, a pairwise comparison might not be an option for this case.

Our robustness score evaluates OODD approaches over several optimization methods, however, someone may decide to treat optimizer selection as hyperparameter tuning. This can be performed in two ways: i) We can tune it to improve classification performance; the improvement is typically small. We could not observe in our experiments that there is a strong correlation between the slightly improved classification accuracy and OODD performance. This tuning will not affect the issue at hand and our robustness score. ii) We can tune the optimization for a specific OODD approach until we get the best OODD performance. This might cause the model to have a bias toward some OOD datasets; however, it can be helpful when a user has already selected an OODD approach and wants to get the maximum OODD performance out of it. This case can bring up the pairwise comparison idea (mentioned in the previous paragraph).

We also attempted to find a correlation between the final minima and OOD performance. We looked at published articles in the field of optimization methods to see if there are any insights about the properties of the final minima that we could apply. Unfortunately, we could not find any relevant work. We also tried to apply statistical tests to feature spaces of the final minima to understand trends or patterns, but we could not see any patterns. Some of our statistical tests failed due to the sparsity of neurons.

Our robustness score is a statistically rigorous formulation to decide which approach is robust in general w.r.t optimizers. An OODD approach can have the best detection performance for a specific combination of optimizer, ID and OOD data. If we compare detection approaches by selecting the best optimizer for each approach, the comparison would be valid only for that specific combination of ID and OOD data. Our robustness score accounts for all these combinations and shows which approach performs well in general. It also considers the tradeoff between mean and variance for better comparison.

Finally, in our experiments, we trained limited number of ID and OOD models with different optimization methods and fixed parameters (e.g., learning rate). It is possible that changing such parameters may affect some of the results; however, the general approach to incorporate varying optimization methods in OODD performance assessment remains unaffected.

\section{CONCLUSION AND FUTURE WORK}
In this paper, we concentrated on the robustness of OODD approaches for deep classifiers trained with different optimization methods and achieving the same classification accuracy. OODD approaches can have varying detection performance when they are applied to such classifiers. Thus, a comparison of OODD approaches under one optimization method cannot result in sound comparison because one OODD approach can outperform others under such an optimization method.

We proposed a robustness score which takes into account variations in OODD metrics due to optimization methods. We also proposed a way to find best-performing optimization methods for OODD approaches. We compared several OODD approaches based on our proposed robustness score and find their best-performing optimization methods. The result reveals that there is not a winner among several OODD approaches investigated in this paper when the effect of optimization methods is considered. It also shows that it is feasible to have better OODD for a specific OODD approach just by using the right optimization method for a given ID dataset.

As future work, it is interesting to investigate whether or not the issue indicated in this paper stands for other OODD approaches that need ID models to be retrained. We also want to investigate the effect of optimization methods' parameters on the performance of OODD approaches.

\small
\bibliography{main}
\bibliographystyle{unsrtnat} 

\newpage

\section*{Appendix}
\begin{table}[h]
\centering
\caption{Test classification accuracy of a CNN for MNIST under different optimization methods.}
    \begin{tabular}{lc}
        \toprule
        Optimizer & accuracy \\
        \bottomrule
        
        Adam      & \multicolumn{1}{l}{\textit{0.9995}}  \\
        RMSprop   & \multicolumn{1}{l}{\textit{0.9984}} \\
        Adamax    & \multicolumn{1}{l}{\textit{0.9998}} \\
        Nadam     & \multicolumn{1}{l}{\textit{0.9994}}\\ 
        SGD       & \multicolumn{1}{l}{\textit{0.9983}} \\
        Adagrad   & \multicolumn{1}{l}{\textit{0.9992}} \\
        Adadelta  & \multicolumn{1}{l}{\textit{0.9984}}\\
        
    \end{tabular}
    
    \label{table_mnist_try02_accuracies}
\end{table}

\begin{figure}[h]
    \centering
    \includegraphics[width=7.5cm]{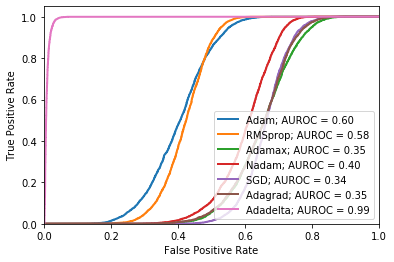}
    \caption{AUROC (based on the MD approach) of models in Table ~\ref{table_mnist_try02_accuracies} when the OOD dataset is uniform noise. }  
    \label{mnist_accuracy_md_unifrom}
\end{figure}

\begin{figure*}
\centering
\subcaptionbox{\label{mnist_md_max-softmax_optimizers_comp:ma}}{\includegraphics[width=6cm]{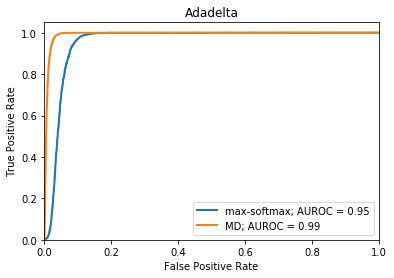}}
\subcaptionbox{\label{mnist_md_max-softmax_optimizers_comp:mb}}{\includegraphics[width=6cm]{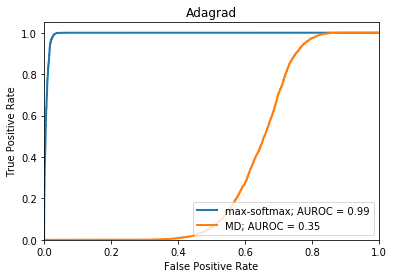}}

\caption{AUROC for detecting uniform noise (as OOD) by max-softmax and MD over the test set of MNIST (as ID). a) It shows AUROC for these two OODD approaches under the Adadelta optimizer. b) It shows AUROC for these two approaches under the Adamax optimizer. As seen, there is not a winner.}  
\label{mnist_md_max-softmax_optimizers_comp}
\end{figure*}

\begin{table*}
  \caption{Values of parameters for optimization methods used in our experiments. The parameters are based on the default values used in Keras.}
  \label{opt_name_values_para}
  \centering
  \begin{tabular}{lc}
  \toprule
  Methods  & Parameters and their values \\
  \bottomrule
  
  Adam      & \multicolumn{1}{l}{\textit{ lr=0.001, beta\_1=0.9, beta\_2=0.999, epsilon=None, decay=0.0, amsgrad=False}}  \\
  RMSprop   & \multicolumn{1}{l}{\textit{lr=0.001, rho=0.9, epsilon=None, decay=0.0}} \\
  Adamax    & \multicolumn{1}{l}{\textit{lr=0.001, beta\_1=0.9, beta\_2=0.999, epsilon=None, decay=0.0}} \\
  Nadam     & \multicolumn{1}{l}{\textit{lr=0.001, beta\_1=0.9, beta\_2=0.999, epsilon=None, schedule\_decay=0.004}}\\ 
  SGD       & \multicolumn{1}{l}{\textit{lr=0.01, momentum=0.0, decay=0.0, nesterov=False}} \\
  Adagrad   & \multicolumn{1}{l}{\textit{lr=0.01, epsilon=None, decay=0.0}} \\
  Adadelta  & \multicolumn{1}{l}{\textit{lr=0.1, rho=0.95, epsilon=None, decay=0.0}}\\
  
  \bottomrule

  \end{tabular}
  \label{optmization_parameters}
\end{table*}

\begin{table*}
    \centering
    \caption{$\mu^{\zeta,t}$ and $\mathrm{Var}^{\zeta,t}$ when $\zeta=($Fashion-MNIST, MNIST, max-softmax$)$ and $t = $Adam. $\downarrow$ indicates that lower values are better; $\uparrow$ indicates that higher values are better.}
    \begin{tabular}{|l|l|l|l|l|l|}
    \hline

    \multicolumn{1}{|c|}{\begin{tabular}[c]{@{}c@{}}Opt.\\ method\end{tabular}} &  
    \multicolumn{1}{|c|}{\begin{tabular}[c]{@{}c@{}}FPR at\\ 95\% TPR $\downarrow$ \end{tabular}}&
    \multicolumn{1}{c|}{\begin{tabular}[c]{@{}c@{}}Detection \\ error $\downarrow$    \end{tabular}} & 
    \multicolumn{1}{|c|}{\begin{tabular}[c]{@{}c@{}}AUROC \\ $\uparrow$ \end{tabular}}&
    \multicolumn{1}{c|}{\begin{tabular}[c]{@{}c@{}}AUPR\\Out \\ $\uparrow$ \end{tabular}} & 
    \multicolumn{1}{|c|}{\begin{tabular}[c]{@{}c@{}}AUPR\\In \\ $\uparrow$ \end{tabular}} \\ \hline
 Adam & 77.356 & 41.465 & 64.109 & 67.773 & 61.175 \\
 Adam & 69.035 & 37.01 & 74.141 & 76.379 & 72.005 \\
 Adam & 68.165 & 36.67 & 70.267 & 74.257 & 65.598 \\
 Adam & 71.815 & 50.0 & 68.161 & 71.84 & 65.366 \\
 Adam & 70.663 & 38.435 & 67.484 & 71.606 & 63.916 \\
\hline
 $\mu^{\zeta,t} | \mathrm{Var}^{\zeta,t}$  & 71.407 | 10.45 & 40.716 | 24.412 & 68.832 | 10.964 & 72.371  | 8.326 & 65.612 | 12.699 \\ \hline
\end{tabular}
\label{fmnist_mnist_r_1}
\end{table*}
\begin{table*}
    \centering
    \caption{$\mu^{\zeta}$ and $\mathrm{Var}^{\zeta}$ when $\zeta=($Fashion-MNIST, MNIST, max-softmax$)$ and $\mathcal{T} =\{$Adam, RMSprop, Adamax, Nadam, SGD, Adagrad, Adadelta$\}$. $\downarrow$ indicates that lower values are better; $\uparrow$ indicates that higher values are better.}
    \begin{tabular}{|l|l|l|l|l|l|}
    \hline

    \multicolumn{1}{|c|}{\begin{tabular}[c]{@{}c@{}}Opt. \\ method \end{tabular}} &  
    \multicolumn{1}{|c|}{\begin{tabular}[c]{@{}c@{}}FPR at\\ 95\% TPR \\ $\mu \downarrow | \mathrm{Var} \downarrow$ \end{tabular}}&
    \multicolumn{1}{c|}{\begin{tabular}[c]{@{}c@{}}Detection \\ error \\ $\mu \downarrow | \mathrm{Var}\downarrow$    \end{tabular}} & 
    \multicolumn{1}{|c|}{\begin{tabular}[c]{@{}c@{}}AUROC \\ $\mu \uparrow | \mathrm{Var} \downarrow$ \end{tabular}}&
    \multicolumn{1}{c|}{\begin{tabular}[c]{@{}c@{}}AUPR\\Out \\ $\mu\uparrow | \mathrm{Var} \downarrow$ \end{tabular}} & 
    \multicolumn{1}{|c|}{\begin{tabular}[c]{@{}c@{}}AUPR\\In \\ $\mu\uparrow | \mathrm{Var} \downarrow$ \end{tabular}} \\ \hline
 Adam & 71.407 | 10.45 & 40.716 | 24.412 & 68.832 | 10.964 & 72.371 | 8.326 & 65.612 | 12.699 \\
 RMSprop & 74.546 | 80.274 & 42.493 | 40.274 & 67.514 | 45.959 & 71.179 | 34.877 & 64.166 | 37.597 \\
 Adamax & 82.092 | 82.584 & 45.235 | 34.119 & 64.323 | 54.821 & 67.134 | 41.227 & 62.532 | 55.91 \\
 Nadam & 80.101 | 43.682 & 45.573 | 29.507 & 65.14 | 15.893 & 67.955 | 15.701 & 62.74 | 13.509 \\
 SGD & 76.609 | 29.081 & 46.088 | 28.054 & 67.636 | 15.044 & 70.545 | 13.598 & 64.531 | 15.677 \\
 Adagrad & 87.011 | 26.786 & 50.0 | 0.0 & 60.492 | 24.628 & 63.956 | 14.025 & 58.995 | 17.294 \\
 Adadelta & 74.469 | 53.346 & 42.422 | 42.281 & 68.537 | 8.124 & 71.838 | 10.378 & 65.591 | 5.486 \\
\hline
 $\mu^{\zeta} | \mathrm{Var}^{\zeta}$ &77.387 | 67.105 &49.993 | 0.08 &66.469 | 27.413 &69.489 | 25.281 &63.71 | 22.495  \\\hline
\end{tabular}
\label{fmnist_mnist_r_2}
\end{table*}
\begin{table*}
    \centering
    \caption{$\mu^{\zeta}$ and $\mathrm{Var}^{\zeta}$ when $\zeta=($Fashion-MNIST, MNIST, $m )$ and $m \in \{$max-softmax, ODIN,MD, Entropy, Margin, MC-D, MI $\}$. $\downarrow$ indicates that lower values are better; $\uparrow$ indicates that higher values are better.}
    \begin{tabular}{|l|l|l|l|l|l|}
   \hline

    \multicolumn{1}{|c|}{\begin{tabular}[c]{@{}c@{}}OODD \\ approaches \end{tabular}} &  
    \multicolumn{1}{|c|}{\begin{tabular}[c]{@{}c@{}}FPR at\\ 95\% TPR \\ $\mu \downarrow | \mathrm{Var} \downarrow$ \end{tabular}}&
    \multicolumn{1}{c|}{\begin{tabular}[c]{@{}c@{}}Detection \\ error \\ $\mu \downarrow | \mathrm{Var}\downarrow$    \end{tabular}} & 
    \multicolumn{1}{|c|}{\begin{tabular}[c]{@{}c@{}}AUROC \\ $\mu \uparrow | \mathrm{Var} \downarrow$ \end{tabular}}&
    \multicolumn{1}{c|}{\begin{tabular}[c]{@{}c@{}}AUPR\\Out \\ $\mu\uparrow | \mathrm{Var} \downarrow$ \end{tabular}} & 
    \multicolumn{1}{|c|}{\begin{tabular}[c]{@{}c@{}}AUPR\\In \\ $\mu\uparrow | \mathrm{Var} \downarrow$ \end{tabular}} \\ \hline
 max-softmax &77.387 | 67.105 &49.993 | 0.08 &66.469 | 27.413 &69.489 | 25.281 &63.71 | 22.495 \\
 ODIN &75.435 | 71.953 &40.22 | 18.058 &74.588 | 41.227 &73.864 | 49.736 &74.426 | 36.183 \\
 MD &9.877 | 9.661 &7.385 | 2.576 &97.946 | 0.488 &98.243 | 0.359 &97.57 | 0.708 \\
 Entropy &76.023 | 48.192 &40.513 | 12.051 &67.435 | 28.977 &69.975 | 32.341 &66.06 | 27.431 \\
 Margin &76.836 | 62.898 &42.105 | 30.326 &66.139 | 25.828 &69.361 | 25.381 &62.046 | 17.658 \\
 MC-D &55.175 | 71.958 &30.08 | 18.0 &82.466 | 11.97 &83.791 | 14.493 &81.062 | 12.711 \\
 MI &48.207 | 154.677 &26.596 | 38.667 &92.402 | 8.235 &90.691 | 14.23 &93.955 | 4.814  \\\hline
\end{tabular}
\label{fmnist_mnist_r_3}
\end{table*}

\end{document}